\documentclass[letterpaper, 10 pt, conference]{ieeeconf}  

\IEEEoverridecommandlockouts                              

\title{\LARGE \bf Dominating Set Database Selection for Visual Place Recognition}
\usepackage{color, colortbl}
\usepackage[noadjust]{cite}
\usepackage{graphicx}

\usepackage{amsmath}
\usepackage{multirow}
\usepackage{amsmath} 
\usepackage{amssymb}  
\usepackage{diagbox}
\usepackage{graphicx}
\usepackage{algorithm2e}
\usepackage{balance}
\usepackage{caption}
\usepackage{subcaption}
\usepackage{multirow}
\usepackage{booktabs}
\usepackage[free-standing-units=true]{siunitx}

\usepackage{tabularx}
\usepackage{url}

\newcommand\Tstrut{\rule{0pt}{2.3ex}}

\author{Anastasiia Kornilova$^{1*}$, Ivan Moskalenko$^{1,2*}$, Timofei Pushkin$^{1,2}$, Fakhriddin Tojiboev$^{1}$, \\
Rahim Tariverdizadeh$^{1}$, and Gonzalo Ferrer$^{1}$
}

\begin{document}

\twocolumn[{%
\renewcommand\twocolumn[1][]{#1}
\maketitle

\begin{center}
    \centering
    \captionsetup{type=figure}
    \includegraphics[width=0.98\textwidth]{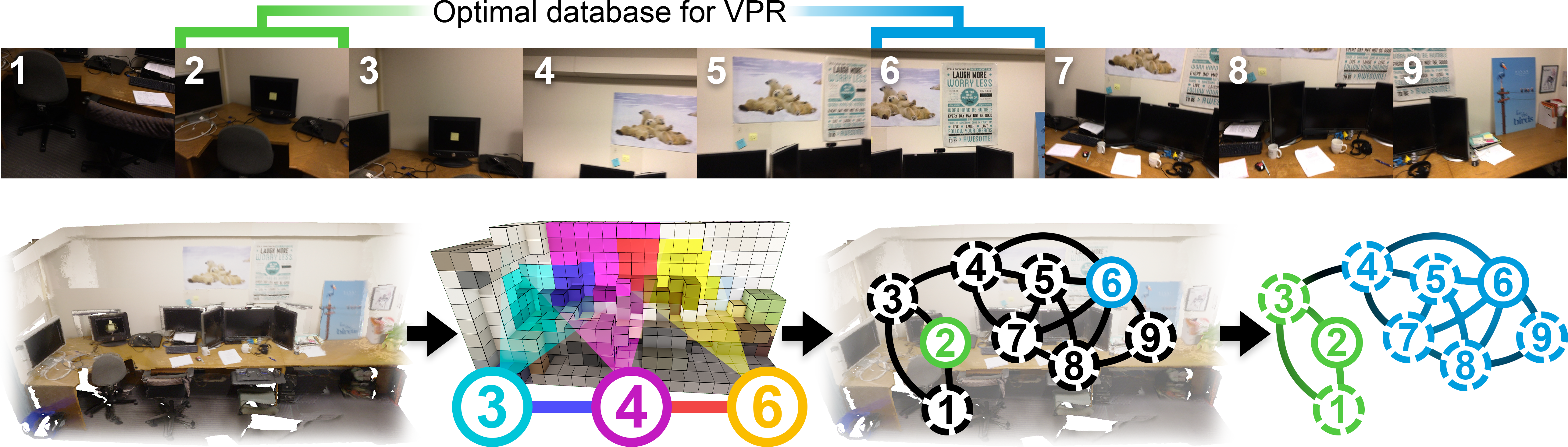}
    \captionof{figure}{Overview of the proposed methodology for building an optimal database for Visual Place Recognition (VPR). \emph{Top:} a sequence of RGBD images obtained from the environment scanning. \emph{Bottom:} (i) a 3D environment map is built from the scanning sequence, (ii) an overlap measure is estimated for each image pair using the spatial overlap in the voxelized map, (iii) the optimal VPR database is built by solving the dominating set problem for a graph where the localized images are vertices connected based on their estimated overlap, (iv) the rest of the images are split into database classes for VPR fine-tuning on the scanned scene.}
    \label{fig:teaser}
\end{center}
}]



\begin{abstract}

This paper introduces a novel approach for creating a visual place recognition (VPR) database for localization in indoor environments from RGBD scanning sequences. The proposed method formulates the problem as a minimization challenge by utilizing a dominating set algorithm applied to a graph constructed from spatial information, referred to as the ``DominatingSet'' algorithm. Experimental results on various datasets, including 7-scenes, BundleFusion, RISEdb, and a specifically recorded sequences in a highly repetitive office setting, demonstrate that our technique significantly reduces database size while maintaining comparable VPR performance to state-of-the-art approaches in challenging environments. Additionally, our solution enables weakly-supervised labeling for all images from the sequences, facilitating the automatic fine-tuning of VPR algorithm to target environment.  Additionally, this paper presents a fully automated pipeline for creating VPR databases from RGBD scanning sequences and introduces a set of metrics for evaluating the performance of VPR databases. The code and released data are available on our web-page~--- \url{https://prime-slam.github.io/place-recognition-db/}.
\end{abstract}

\addtocounter{footnote}{1}
\footnotetext{The authors are with Skolkovo Institute of Science and Technology (Skoltech), Center for AI Technology (CAIT).
         {\tt\small {anastasiia.kornilova, g.ferrer}@skoltech.ru}}
\addtocounter{footnote}{1}
\footnotetext{The authors are with Software Engineering Department, Saint Petersburg State University.
\\
\hspace*{0.3cm}$^*$ Indicates equal contribution.}
\addtocounter{footnote}{-2}


\section{Introduction}

Visual place recognition (VPR) plays a crucial role in solving the localization problem using image data alone. The applications are immense and the Robotics community as well as the Computer Vision community are actively investigating new solutions. A commonly used approach in VPR systems involves two key components: (i) a database consisting of a set of images and their corresponding 3D poses, and (ii) an algorithm that identifies the most similar image in the database to a given query image and estimates its pose relative to the database image. This paper investigates VPR from the perspective of selecting a small subset of images from a sequential stream of sensor-generated data to accurately represent the environment for VPR operation.

Scanning the environment requires a collection of visual data and its corresponding locations, that in indoors is usually solved by SLAM~\cite{campos2021orb, zhang2021survey} or Bundle Adjustment~\cite{schonberger2016structure} algorithms. However, a challenge arises after the scanning process in determining which images should be included in the database. The volume of observations obtained after scanning is often massive, with thousands of observations or more, and highly redundant due to the sequential capture of the scene during movement. To optimize computational and memory resources, especially in low-computational and embedded devices, the database should be compact in size. Furthermore, it should provide sufficient data diversity and coverage of the entire scene to facilitate accurate VPR.

VPR is particularly applicable in indoor environments where the use of Global Navigation Satellite Systems (GNSS) is unavailable, and other global localization equipment such as radio or Wi-Fi beams pose challenges in terms of maintenance and cost. Over the past years, significant efforts have been dedicated to developing robust place recognition algorithms using available benchmarks and datasets, such as Mapillary~\cite{neuhold2017mapillary}, Nordland~\cite{sunderhauf2013we}, Pittsburgh~\cite{torii2013visual}, Tokyo24/7~\cite{torii201524}, and RobotCar Seasons~\cite{maddern20171, toft2020long}. However, these datasets primarily focus on outdoor environments and already provide pre-selected databases for VPR methods.

In this work, we propose an approach for creating VPR databases in indoor environments by utilizing RGBD information and selecting an optimal subset of images from the scanning sequence. To achieve this, we formulate a minimization problem for VPR databases and propose a solution based on the "dominating set" algorithm~\cite{garey1979computers} applied to graphs and spatial information.  A byproduct of this solution is the clusterization of localized images around the selected one, and its application to related tasks such as automatic creation of weakly-supervised images of the same ``place'' for fine-tuning neural algorithms to a specific environment. We demonstrate that our technique, which reduces database size, produces compact databases while maintaining comparable visual localization quality to state-of-the-art approaches in challenging environments.

The main contributions of the paper are as follows:
\begin{itemize}
    \item a formal definition of an optimal VPR database in terms of size and coverage, along with an approach for its calculation;
    \item a fully automated pipeline for VPR database creation from an RGBD scanning sequence, which is made publicly available as a library;
    \item a fully automated end-to-end methodology, from sequence scanning to VPR fine-tuning for a specific environment.
\end{itemize}

\section{Related Work}

VPR algorithms address the task of identifying the most suitable image match from a \textit{database} that represents a given environment, based on a so-called \textit{query} observation captured within the same environment. This problem, known as {\em Information Retrieval}, is common in various fields such as Natural Language Processing, Computer Vision, and Robotics.

To enable VPR, it is necessary to define an \textit{image descriptor} and a \textit{similarity measure} between pairs of descriptors (images). A classic approach involves calculating local image features and aggregating them into a global image descriptor using techniques like bag of words ~\cite{galvez2012bags}, VLAD ~\cite{jegou2010aggregating, mironicua2016modified}, or differentiable NetVLAD ~\cite{arandjelovic2016netvlad, hausler2021patch}. Local image features can be computed using handcrafted algorithms~\cite{lowe2004distinctive, bay2006surf, rublee2011orb} or by employing learnable methods for keypoint detection and description~\cite{yi2016lift, detone2018superpoint, barroso2019key}.

The calculation of the global image descriptor is an active area of research, with algorithms achieving remarkable performance. For instance, Hloc~\cite{sarlin2019coarse} learn to predict both global and local features simultaneously, while CosPlace~\cite{berton2022rethinking} provides a learned global descriptor without intermediate local feature aggregation. In this study, we utilize state-of-the-art global image descriptors as a tool for VPR.  It can be argued that only with the utilization of these advanced VPR methods can we significantly reduce the dataset size, as proposed in this paper.

Existing VPR approaches predominantly concentrate on urban outdoor environments and datasets~\cite{neuhold2017mapillary, sunderhauf2013we, torii2013visual, torii201524, maddern20171, toft2020long}. This emphasis arises from the availability of training data in outdoor scenarios, where reference poses can be directly obtained using GNSS technologies. Even with GNSS sensor errors of up to a meter, it is still possible to establish accurate associations for image correspondences. The creators of these datasets already provide a comprehensive database and training correspondences. 

While the majority of research focuses on outdoor environments, there have been limited efforts to address indoor visual place recognition and the associated database creation. The 7-Scenes dataset~\cite{glocker2013real, shotton2013scene} is commonly employed for training and evaluating indoor localization methods. However, it only covers small areas like office rooms and apartment sections. Other datasets, such as TUM-RGBD~\cite{sturm2012benchmark}, ScanNet~\cite{dai2017scannet}, and BundleFusion~\cite{dai2017bundlefusion}, are used for RGB-D SLAM and also offer the opportunity to test visual localization methods. However, they are limited in terms of scene coverage and variations in lighting conditions. In contrast, the RISEdb dataset~\cite{sanchez2020rise} includes large-scale scenes that encompass entire floors of various types of buildings. Additionally, this dataset provides data captured under varying lighting conditions at different times of the day.

The creation of a map database is also indirectly addressed in SLAM pipelines through a process known as \textit{keyframe selection} and used as a basis for SLAM-graph optimization and loop closure. Various techniques have been employed in different systems to determine keyframes based on different criteria. ORB-SLAM~\cite{campos2021orb} and its subsequent versions~\cite{li2021rgb, pumarola2017pl} utilize bag-of-words representation combined with local features. In these approaches, an image is considered a keyframe if it observes a significant number of new local features compared to the existing map. BundleFusion~\cite{dai2017bundlefusion} divides the data stream into chunks, each containing an equal number of RGBD images. From each chunk, a single keyframe is selected for further processing. Das et al.~\cite{das2015entropy} propose two methods based on image entropy to estimate whether a keyframe will contribute to map improvement or not. These approaches evaluate the information content of the keyframe relative to the existing map. Alonso et al.~\cite{alonso2019enhancing} exploit image quality criteria such as blurriness and brightness, as well as semantic content criteria based on a CNN called MiniNet, to assess the suitability of keyframes for enhancing the map. Sheng et al.~\cite{sheng2019unsupervised} introduce a joint learning approach for keyframe detection and visual odometry, where the system learns to identify keyframes that are informative for both tasks. iMap~\cite{sucar2021imap} and NICE-SLAM~\cite{zhu2022nice} adopt a depth overlap criterion with respect to the map to measure the amount of new information that a potential keyframe can contribute to the map. Mainly, the process of keyframe selection primarily targets the internal workings of the SLAM algorithm, which could potentially restrict its effectiveness for the global visual place recognition task when compared to the technique of selecting general subsets of images that were captured in the environment.

\section{Methodology}

The proposed methodology for creating an optimal database for visual place recognition (VPR) is illustrated in Fig.~\ref{fig:teaser}. The algorithm takes color images, corresponding depth images, and their poses as input. The poses can be obtained through classical state estimation techniques like SLAM-based solutions or beacon systems. Using this information, a 3D map of the environment is generated. Next, the 3D map is divided into voxels, and the spatial overlap between pairs of images is calculated by determining the intersection between the voxel sets of the images. This overlap information enables the construction of a graph that represents the connections between the images based on their calculated overlaps. Finally, an optimal database for VPR is obtained by identifying a dominating set within this graph. Optionally, this methodology allows for the division of the remaining images from the scanning sequence into database classes, which can be utilized for training or fine-tuning VPR algorithms.

\subsection{Problem formulation for optimal database}

Consider a given set of color images that capture the environment, denoted as $C = {c_1, \ldots, c_N}$. For any two color images $c_i$ and $c_j$, an \textit{overlap measure} $s(c_i, c_j) \in [0, 1]$ is defined to quantify the extent to which the view scopes of the images intersect. To assess the coverage of a subset of color images $\widetilde{C} \subset C$, a \textit{coverage loss} function $f(\cdot)$ is introduced. This function quantifies the number of images covered by the subset and is defined as follows:

\begin{align}
f_C(\widetilde C) = \sum_{c \in C}
\begin{cases}
      0 & \text{if $\exists \widetilde c \in \widetilde C$ s.t. $s(c, \widetilde c) > \mu$ } \\
      1 & \text{else}
\end{cases}
\end{align}
where $\mu$ is an overlap threshold. Our objective is to identify a subset $\widetilde{C}$ of the minimum size that ensures coverage of all frames in $C$:

\begin{equation}
    C_{db} = \min_{\substack{\widetilde C \subset C \\ f_C(\widetilde C) = 0}} | \widetilde C|.
\end{equation}

In this context, we define $C_{db}$ as an \textbf{optimal image database}, which serves as a condensed representation of the color information obtained from digitizing the environment.

When the overlap measure between images satisfies symmetry, i.e., $s(c, \widetilde{c}) = s(\widetilde{c}, c)$, the problem formulation resembles the minimum dominating set problem in graph theory. In this analogy, the images are represented as vertices in a graph, where an edge exists between two vertices if their overlap measure exceeds a given threshold. The original dominating set problem aims to identify the smallest subset of vertices in the graph such that every vertex either belongs to the subset or is adjacent to a vertex within it.

\subsection{Spatial overlap measure}

In our methodology, we propose utilizing spatial information obtained from the depth camera to estimate the overlap between two images. This approach provides an advantage over an alternative approach that relies solely on color information, such as using local features and matches between them. The use of color information alone can lead to incorrect edges being generated, particularly in situations where different locations exhibit similar textures and patterns, resulting in visual aliasing.

With access to depth information for each color frame, we can construct a 3D map of the environment using image poses. This 3D map can be represented as a set of voxels, denoted as $V = \{v_1, \ldots, v_M\}$. The sequence of color images, $C$, is associated with the voxels through the set $D = \{d_1, \ldots, d_N\}$, where $d_i = \{v_{i_1}, \ldots, v_{i_k}\}$ represents a subset of voxels observed in frame $c_i$. By considering these subsets of voxels, we can define an overlap measure, denoted as $s(d_i, d_j)$, based on the intersections of voxel sets. This overlap measure provides a more accurate estimation of the degree of overlap compared to considering only color information. In our analysis, our objective is to identify a subset $\widetilde{D} \subset D$ that satisfies the following criterion:

\begin{align}
    D_{db} = \min_{\substack{\widetilde D \subset D \\ 
    f_D(\widetilde D) = 0}} 
    | \widetilde D|.
\end{align}

\subsection{Overlap measure}

\begin{figure}
    \centering
    \includegraphics[width=0.8\linewidth]{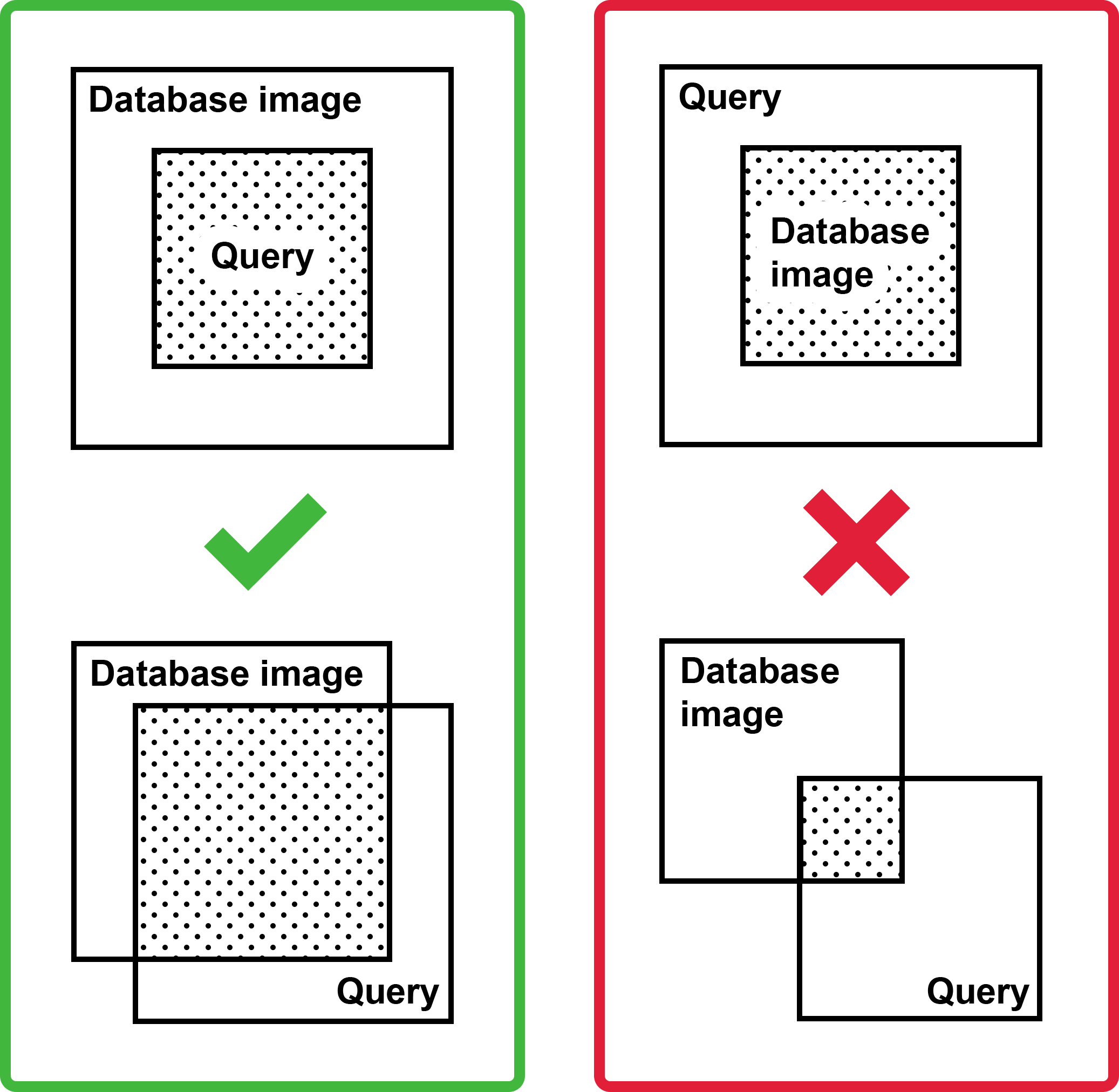}
    \caption{Examples showcasing different overlaps between a query and a database image, interpreted as good or bad. \emph{Left}: the database image covers a significant portion of the query image. \emph{Right}: the query image is not well covered by the database image.}
    \label{fig:overlap_measure}
\end{figure}

To establish an appropriate overlap measure, let us examine the examples illustrated in Fig.~\ref{fig:overlap_measure}. A database image can be considered \textit{good} in relation to a query image if it covers a significant part of the query image. Conversely, a database image can be deemed \textit{bad} if it covers only a relatively small part of the query image. It is important to note that this principle is not symmetric. A query image may occupy a minor portion of a database image, yet the database image may still provide sufficient coverage for the query. Formally, the overlap measure for voxel sets can be defined as follows:

\begin{align}
\label{eq:overlap}
    s(d_q, d_{db}) = \frac{| d_q \cap d_{db} |}{| d_q|},
\end{align}
where $|\cdot|$ is set cardinality and $d_q$ and $d_{db}$ are voxel sets corresponding to the query and database images respectively.

As mentioned previously, in the case where the overlap measure exhibits symmetry, the minimization problem can be addressed using dominating set algorithms and existing solvers~\cite{hagberg2008exploring}. In our study, we propose employing the intersection over union (IoU) as the overlap measure for voxel sets:

\begin{align}
    s_{IoU}(d_q, d_{db}) = \frac{| d_q \cap d_{db} |}{| d_q \cup d_{db}|}
\end{align}

This metric, being symmetric, imposes stricter constraints on the graphsince:

\begin{align}
    s_{IoU}(d_q, d_{db}) \leq s(d_q, d_{db}) 
\end{align}

The inclusion of stricter constraints leads to an increased number of edges in the graph, which, in turn, may result in a larger database volume when applying the dominating set solution compared to the original problem formulation.

\subsection{Graph processing}

To construct the graph, it is necessary to determine the overlap between every pair of images. However, in cases where the scanning sequence is extensive, analyzing the intersection between $\frac{N(N-1)}{2}$ pairs of voxel sets can produce computational overhead. Moreover, many pairs of voxel sets may not have any intersection. To address these challenges and optimize the process, we propose the following algorithm:

\begin{enumerate}
    \item associate each voxel in the map with the indices of the frames that cover it;
    \item generate pairs of frames for each voxel, encompassing all possible combinations;
    \item for each pair, calculate the number of voxels where the same pair of frames occurs. This count reflects the size of the corresponding intersection;
    \item finally, compute the Intersection over Union (IoU) for the constructed pairs using the previously calculated intersection sizes. 
\end{enumerate}

By employing this approach, only the voxel sets with non-empty intersections need to be considered, leading to a more efficient graph construction process.
\section{Experimental results}

In this section, we present a comparative analysis of the quality of the VPR database constructed using our method in contrast to other existing strategies for database creation. Our evaluation focuses on two primary test cases. The first test case involves indoor localization sequences taken from the 7-scenes and BundleFusion datasets. The second test case expands upon this evaluation by conducting experiments in more challenging and large-scale environments, specifically the RISEdb dataset (with lightning changes) and our self-curated Skoltech Campus dataset. Furthermore, we showcase the performance of cutting-edge VPR algorithms that have been adapted to operate with the database generated using our methodology.

\subsection{Datasets}

To assess the quality of a database and the adapted VPR algorithm,  considered datasets should resemble real-world scenarios. This entails dividing the dataset into a ``scanning'' sequence for database creation and VPR algorithm adaptation, and a separate ``test'' sequence for evaluating place recognition quality. To fulfill these requirements, our evaluation includes the two largest scenes from each considered dataset: 7-scenes dataset (RedKitchen and Office)~\cite{glocker2013real, shotton2013scene}, BundleFusion (office0 and office1)~\cite{dai2017bundlefusion}, RISEdb (as3ml and auditorium)~\cite{sanchez2020rise}. For the 7-scenes dataset, the sequences labeled as ``train'' in the original dataset are designated as the scanning sequence, while the remaining sequences are utilized for testing. As for the BundleFusion dataset, we manually partitioned the sequence into a dedicated scanning route and a distinct test route. In the case of the RISEdb dataset, where multiple sequences exist for each scene, we selected the largest sequence as the scanning route and the second largest sequence as the test route. As RISEdb does not provide depth images but only a 3D map, we projected the map onto the corresponding color frames in order to obtain depth information.

\begin{figure*}[t!]
    \centering
    \includegraphics[width=\textwidth]{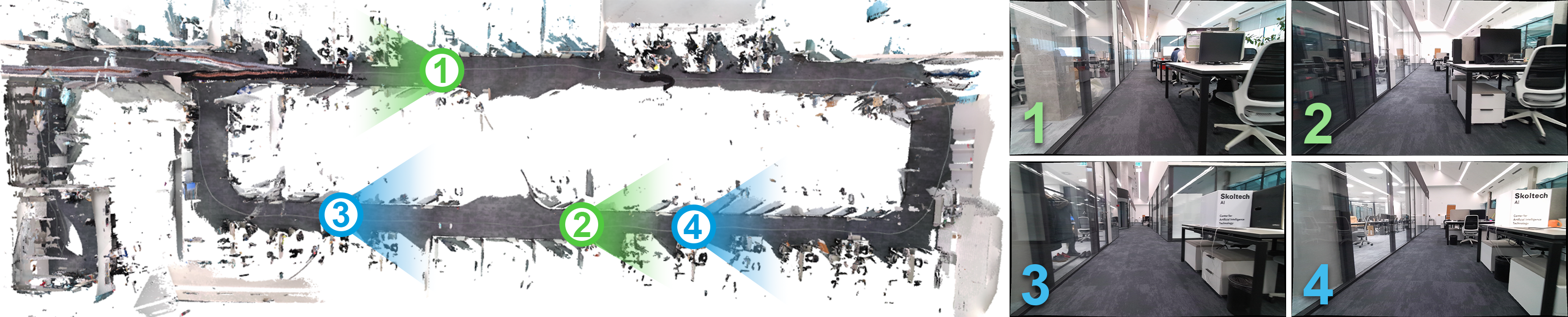}
    \caption{Examples of different scenes captured in Skoltech campus having similar structures. \emph{Left:} the map of the Skoltech campus sequence. \emph{Right:} pairs of images that are very similar visually but were captured in different locations.}
    \label{fig:sk_map}
\end{figure*}


Furthermore, we have captured a series of sequences within the Skoltech campus using an Azure Kinect DK sensor. This particular environment presents a challenge for VPR algorithms in indoor settings due to the presence of recurring structures with similar designs, such as desks, walls, and doors, resulting in increased visual ambiguity. Illustrations depicting some instances of visual ambiguity in this scene can be seen in Fig.~\ref{fig:sk_map}. To construct a comprehensive 3D map, we utilize the trajectory generated by the RGBD ORB-SLAM algorithm~\cite{campos2021orb}. However, it should be noted that certain sensors may not provide complete depth coverage for each color frame. To address this limitation, we have implemented a mechanism to extend the depth coverage using 3D map reprojection to the precise frame.

\subsection{Database size and reduction rate}

\begin{table}[]
\caption{\label{tab:db_size} Statistics on the scanning sequence size and the size of the resulting databases}
\begin{tabular}{lcccc}
\hline
\multirow{5}{*}{Dataset}              & \multirow{5}{*}{Sequence size} & \multicolumn{3}{c}{Overlap threshold}\Tstrut \\
                                      &                                & 0.1   & 0.3  & 0.5                           \\ \cline{3-5} 
                                      &                                & \multicolumn{3}{c}{Database size}\Tstrut     \\
                                      &                                & \multicolumn{3}{c}{Reduction rate}           \\
                                      &                                & \multicolumn{3}{c}{Spatial coverage}         \\ \hline
\multirow{3}{*}{7-Scenes Office}      & \multirow{3}{*}{4800}          & 5     & 15   & 80\Tstrut                     \\
                                      &                                & x960  & x320 & x60                           \\
                                      &                                & 43\%  & 56\% & 75\%                          \\ \hline
\multirow{3}{*}{7-Scenes RedKitchen}  & \multirow{3}{*}{5600}          & 4     & 25   & 97\Tstrut                     \\
                                      &                                & x1400 & x224 & x58                           \\
                                      &                                & 24\%  & 54\% & 68\%                          \\ \hline
\multirow{3}{*}{BundleFusion office0} & \multirow{3}{*}{3547}          & 14    & 33   & 94\Tstrut                     \\
                                      &                                & x253  & x108 & x38                            \\
                                      &                                & 61\%  & 72\% & 82\%                           \\ \hline
\multirow{3}{*}{BundleFusion office1} & \multirow{3}{*}{3622}          & 12    & 30   & 86\Tstrut                     \\
                                      &                                & x301  & x120 & x42                            \\
                                      &                                & 47\%  & 65\% & 75\%                           \\ \hline
\multirow{3}{*}{Sk campus}            & \multirow{3}{*}{6849}          & 20    & 46   & 105\Tstrut                    \\
                                      &                                & x342  & x148 & x105                           \\
                                      &                                & 50\%  & 65\% & 72\%                           \\ \hline
\multirow{3}{*}{RISEdb as3ml}            & \multirow{3}{*}{10952}          & 102    & 363   & 845\Tstrut                    \\
                                      &                                & x107  & x30 & x13                           \\
                                      &                                & 74\%  & 92\% & 95\%                           \\ \hline
\multirow{3}{*}{RISEdb auditorium}            & \multirow{3}{*}{19278}          & 94    & 307   & 730\Tstrut                    \\
                                      &                                & x205  & x63 & x26                           \\
                                      &                                & 69\%  & 90\% & 96\%                           \\ \hline
\end{tabular}
\end{table}

Our database creation algorithm incorporates two hyperparameters: voxel size for map voxelization and overlap threshold for graph processing. In our experiments, we set the voxel size to 0.3 m, which provides a reasonable approximation of the spatial characteristics of indoor environments captured by RGBD sensors. We consider three overlap thresholds for the Intersection over Union (IoU) measure: 0.1, 0.3, and 0.5. Table 1 presents statistics on the size of the scanning sequence and the resulting size of the constructed database for different overlap threshold values. The table also includes information on the reduction rate and the percentage of spatial coverage achieved by the database. The spatial coverage is quantified as the percentage of map voxels occupied by the voxels covered by the database.

The findings of our study indicate that our proposed methodology yields a significant reduction in the size of the scanning sequence. The extent of this reduction is contingent upon the speed and comprehensiveness of the environmental data recording process. For instance, when the recording process is slow, the reduction rate is found to be the highest, as observed in the 7scenes dataset. It is important to emphasize that the spatial coverage of certain constructed databases is relatively limited, encompassing only 25-40\% of the voxels. Nevertheless, the utilization of invariance in our approach ensures that the selected frames for the database exhibit adequate overlap with other images, thereby covering the remaining voxels that have not been addressed. Additionally, it is noteworthy that the spatial overlap for the two largest sequences in the RISEdb dataset remains largely unchanged when the threshold is adjusted from 0.3 to 0.5. This observation indicates that the saturation point of information has been reached.

\subsection{Visual place recognition}

In this study, we present an evaluation of state-of-the-art visual place recognition algorithms using our methodology for database creation. 

Specifically, we assess the performance of the following methods that have demonstrated superior results on popular visual place recognition (VPR) datasets: NetVLAD~\cite{hausler2021patch}, CosPlace~\cite{berton2022rethinking}, and a combined approach using both NetVLAD and SuperGlue~\cite{sarlin2020superglue}. The combined approach utilizes the top-5 predictions from the NetVLAD algorithm and then applies the SuperGlue method to determine the prediction with the maximum number of matches. The utilization of SuperGlue for all frames is not a viable approach due to its low computational efficiency. This inefficiency stems from the fact that the entire image is processed for every request, rather than solely focusing on the descriptors of the image that can be preprocessed in advance for VPR operation.

\begin{table*}[]
\caption{\label{tab:vpr_eval_little} Recall of Visual Place Recognition algorithms on databases generated by DominatingSet for \textbf{small-scale scenarios}. (f) stands for fine-tuned model.}
\centering
\begin{tabular}{l|ccc|ccc|ccc|ccc}
Dataset &  \multicolumn{3}{c|}{7-Scenes Office} & \multicolumn{3}{c|}{7-Scenes RedKitchen} & \multicolumn{3}{c|}{BundleFusion office0} & \multicolumn{3}{c}{BundleFusion office1} \\ \hline

Overlap threshold & \textbf{0.1} & \textbf{0.3} & \textbf{0.5} & \textbf{0.1} & \textbf{0.3} & \textbf{0.5} & \textbf{0.1} & \textbf{0.3} & \textbf{0.5} & \textbf{0.1} & \textbf{0.3} & \textbf{0.5} \\ \hline

NetVLAD & 63.9 & 81.9 & \textbf{91.3} & 65.0 & 75.6 & \textbf{88.9} & 74.1 & 83.8 & \textbf{88.1} & 70.9 & 87.3 & \textbf{96.0} \\

NetVLAD (f) & \textbf{99.2} & 96.9 & 88.8 & \textbf{95.3} & 93.7 & 93.1 & 84.2 & \textbf{90.8} & 90.1 & 83.9 & \textbf{97.1} & 95.3 \\

NetVLAD (f) + SuperGlue & -- & 98.6 & \textbf{99.0} & -- & 99.1 & \textbf{99.2} & 91.9 & 96.8 & \textbf{98.7} & 87.4 & 99.2 & \textbf{99.7} \\


CosPlace (ResNet-101, 2048) & 64.7 & 83.8 & \textbf{87.8} & 68.2 & 63.1 & \textbf{90.3} & 64.7 & 80.7 & \textbf{90.7} & 62.3 & 79.3 & \textbf{92.6} \\

CosPlace (f) & 88.6 & 97.8 & \textbf{98.4} & 80.8 & 89.3 & \textbf{95.8} & 91.2 & 92.1 & \textbf{95.2} & 82.1 & 97.3 & \textbf{97.4} 
\end{tabular}
\end{table*}

\begin{table*}[]
\caption{\label{tab:vpr_eval_big} Recall of Visual Place Recognition algorithms on databases generated by DominatingSet for \textbf{large-scale scenarios}. (f) stands for fine-tuned model.}
\centering
\begin{tabular}{l|ccc|ccc|ccc}
Dataset & \multicolumn{3}{c|}{Sk campus} & \multicolumn{3}{c|}{RISEdb as3ml} & \multicolumn{3}{c}{RISEdb auditorium} \\ \hline

Overlap threshold & \textbf{0.1} & \textbf{0.3} & \textbf{0.5} & \textbf{0.1} & \textbf{0.3} & \textbf{0.5} & \textbf{0.1} & \textbf{0.3} & \textbf{0.5} \\ \hline

NetVLAD & 56.9 & 80.3 & \textbf{94.5} & 25.6 & 63.5 & \textbf{78.3} & 27.7 & 51.0 & \textbf{70.7} \\

NetVLAD (f) & 88.0 & \textbf{89.1} & 83.4 & 31.4 & 59.4 & \textbf{67.2} & 26.3 & 42.5 & \textbf{50.7} \\

NetVLAD (f) + SuperGlue & 74.9 & 92.3 & \textbf{94.1} & 38.3 & 77.7 & \textbf{82.7} & 40.2 & 59.1 & \textbf{72.9} \\


CosPlace (ResNet-101, 2048) & 63.0 & 80.9 & \textbf{91.9} & 28.3 & 69.1 & \textbf{83.7} & 34.5 & 59.7 & \textbf{80.5} \\

CosPlace (f) & 84.3 & 93.2 & \textbf{95.3} & 40.4 & 85.0 & \textbf{90.5} & 68.3 & 84.7 & \textbf{91.8}

\end{tabular}
\end{table*}

Two types of models for NetVLAD and CosPlace are considered: the original pre-trained models provided by the authors and models that we fine-tuned specifically for each database and scene. To prepare the data, we employ the following pipeline: every fifth frame from the scanning sequence is extracted for the validation set, while the remaining images are used for training. Since the \textit{DominatingSet} algorithm provided class labels, we trained the models on this subset of data until an ``early stop'' criterion was met.

All evaluations are conducted on test sequences that are not included in the database creation pipeline or VPR algorithm fine-tuning process. The evaluation metric used was Recall@1, which is a commonly used metric for VPR tasks. In order to determine whether the query frame and the database frame were correctly matched, we use metric~(\ref{eq:overlap}) with a threshold value equal to 0.3. The results are presented in Tab.~\ref{tab:vpr_eval_little} and Tab.~\ref{tab:vpr_eval_big}, considering small-scale and large-scale scenarios respectively.

First, it can be noticed that the overall localization quality is superior in small-scale scenarios compared to large-scale scenarios, even when employing pre-trained models. This can be attributed to the greater diversity of data captured across the scenes in small-scale scenarios, in contrast to more challenging datasets characterized by repetitive structures and variations in lighting conditions. Secondly, for the creation of databases in small-scale scenarios, an overlap threshold of 0.1 proves to be a suitable choice. This threshold enables a significant reduction in the number of frames within the database while maintaining relatively good localization quality on fine-tuned models. In the case of large-scale applications, a more reasonable choice for the overlap threshold would be 0.3 or 0.5. These values contribute to an improvement in localization quality, albeit with a marginal difference between 0.3 and 0.5. Consequently, memory consumption can be reasonably optimized by utilizing a smaller overlap threshold. Furthermore, the CosPlace method demonstrates the best overall performance. It excels in challenging environments, providing superior quality and comparable performance to other methods in small-scale environments. It is worth noting that the fine-tuning process employed in our methodology enhances the performance of the localization method by tailoring it to the specific environment. However, it should be acknowledged that when using a 0.5 overlap threshold, the fine-tuned NetVLAD exhibits lower quality compared to the original network. This can be attributed to overfitting of the network to specific frames.

In summary, our approach for database creation offers compression capabilities that yield exceptional quality in small-scale environments and comparable performance to state-of-the-art visual localization methods in challenging environments. Also, proposed approach provides a common way to formulate problem of optimal VPR database creation and, in the future, can be applied to any scanning sequences that records spatial information, for example, LiDAR data and outdoor environmetns.

\section{Conclusion}

This paper has presented a method for creating compact and descriptive VPR databases for indoor environments. To do so, our approach defines a formal definition of optimal database and provides an algorithm for its generation from sequential stream of RGBD data. In our experiments, that cover various environments with different conditions, we have demonstrated that scanning sequence can be compressed to compact database while maintaining comparable visual localization quality to state-of-the-art methods. Additionally, we have provided a methodology on top of our method to generate data for fine-tuning data for target environment, that shows improvements with respect to original models on target environments. We have made all contributions, including code and data, public.

\bibliographystyle{IEEEtran} 
\bibliography{bibliography}

\end{document}